\documentclass{article}
\usepackage{spconf,amsmath,amssymb,graphicx}
\usepackage{booktabs}
\usepackage{adjustbox}
\usepackage{pifont}
\usepackage{url}
\newcommand{\cmark}{\ding{51}} % check mark
\newcommand{\xmark}{\ding{55}} % cross mark
\usepackage{xcolor}
\usepackage{lipsum}
\usepackage{enumitem}
\usepackage[numbers]{natbib}

\usepackage{multirow} % Include this in your preamble

\usepackage{xspace}
\def\sigrid{\textsc{Sigrid}\xspace}
\def\sigrids{\textsc{Sigrid}s\xspace}

\def\iou{$\operatorname{IoU}$\xspace}
\def\maxiou{$\operatorname{MaxIoU}$\xspace}

\title{Superpixel Integrated Grids for Fast Image Segmentation}
\twoauthors
 {Jack Roberts}
	{Brown University\\
	Department of Computer Science\\
	Providence, RI, USA}
 {Jeova Farias Sales Rocha Neto}
	{Bowdoin College\\
	Department of Computer Science\\
	Brunswick, ME, USA}
\begin{document}
%\ninept
%
\maketitle
\begin{abstract}
Superpixels have long been used in image simplification to enable more efficient data processing and storage. However, despite their computational potential, their irregular spatial distribution has often forced deep learning approaches to rely on specialized training algorithms and architectures, undermining the original motivation for superpixelations. In this work, we introduce a new superpixel-based data structure, \sigrid (Superpixel-Integrated Grid), as an alternative to full-resolution images in segmentation tasks. By leveraging classical shape descriptors, \sigrid encodes both color and shape information of superpixels while substantially reducing input dimensionality. We evaluate \sigrids on four benchmark datasets using two popular convolutional segmentation architectures. Our results show that, despite compressing the original data, \sigrids not only match but in some cases surpass the performance of pixel-level representations, all while significantly accelerating model training. This demonstrates that \sigrids achieve a favorable balance between accuracy and computational efficiency.
\end{abstract}

\section{Introduction}
\label{sec:intro}

Image segmentation is a core task in computer vision \cite[Chap. 10]{gonzalez2018digital}, enabling applications across diverse fields such as biomedicine \cite{ronneberger2015u} and remote sensing \cite{dadsetan2021superpixels, dang2024joint}, where images must be partitioned into compact, meaningful regions. With the rise of deep neural networks, their ability to extract rich information from visual data \cite[Chap. 12]{gonzalez2018digital} naturally extended to segmentation tasks \cite{minaee2021image}. However, these methods come at the cost of processing and storing vast amounts of data, which in turn requires expensive, high-end hardware.

A classical strategy to address this challenge is to transform pixel-based image representations into superpixel structures \cite{achanta2012slic}. Superpixels group adjacent pixels with similar color or texture into perceptually meaningful regions, simplifying image representation while preserving important boundaries. Owing to their effectiveness for data reduction and abstraction, superpixels soon found applications within deep learning. In some works, they served as multiscale unsupervised feature extractors to enhance CNN-based segmentation algorithms \cite{he2015supercnn, gadde2016superpixel}. In others, they were incorporated directly into the model, for instance as image-dependent pooling layers \cite{kwak2017weakly}, as tokenizers for transformer-based architectures \cite{zhu2023superpixel}, or as the basis for training graph neural networks \cite{dadsetan2021superpixels, avelar2020superpixel}. Finally, several studies have also proposed deep learning models designed specifically to compute superpixelations \cite{yang2020superpixel, jampani2018superpixel, peng2022hers}.

\begin{table}[b]
    \caption{Organizational primitives of image data.}
    \centering
    \small
    \vspace{3pt}
    \setlength{\tabcolsep}{4pt}
    \begin{tabular}{lccc}
        \toprule[1pt]
        \textbf{Property} & \textbf{Pixel} & \textbf{Superpixel} & \textbf{\sigrid} \\
        \midrule[1pt]
        Number of primitives      & Many      & Few      & Few \\
        Structure                 & Regular   & Irregular & Regular \\
        Compatibility with CNNs   & Native    & Challenging & Native \\
        Semantic content per primitive & Low & High & High \\
        \bottomrule[1pt]
    \end{tabular}
    \label{tab:comp}
\end{table}

\begin{figure*}
    \centering
    \includegraphics[width=1\linewidth]{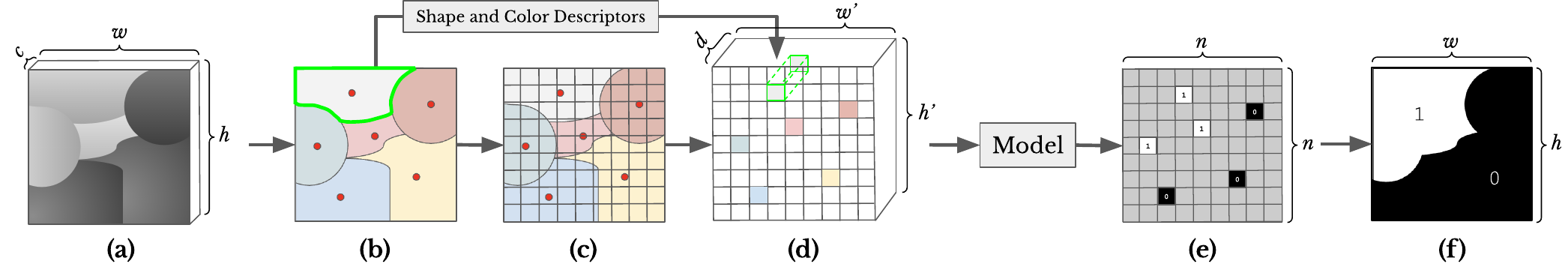}
    \vspace{-20pt}
    \caption{\sigrid construction and model inference for binary segmentation. (a) Original image $I$. (b) Superpixelation $S$ ($K = 6$ here) computed from $I$ and superpixel centers (red dots). (c) An $w'\times h'$ grid superimposed on $S$ where superpixels are assigned to grid cells. (d) Assigned grid cells are then populated with superpixel descriptors. (e) The model classifies each assigned grid cell in one of two classes. (f) Cell classes are converted back to pixel-level classification using the original superpixels.}
    \label{fig:sigrid_construction}
\end{figure*}

Despite these efforts, most approaches still depend on the design of novel and intricate neural architectures to fully exploit superpixel representations. As summarized in Table \ref{tab:comp}, this stems in part from the inherently irregular distribution of superpixels, which complicates their integration into deep learning models built on regular grid structures, such as CNNs. In doing so, these methods end up undermining the very motivation behind superpixels: to simplify not only the image representation but also the subsequent processing.

We introduce a new superpixel-based primitive, \sigrid, designed to replace full-resolution images in segmentation tasks. Our approach efficiently converts dense, pixel-based tensors into compact sparse tensors that encode only superpixel appearance and shape information. When used as input to popular segmentation CNNs, these compressed representations achieve performance comparable to, and in some cases exceeding, networks trained on the original images, while requiring only a fraction of the computational resources. The success of \sigrid arises from the fact that superpixels preserve natural image boundaries, offering a strong structural prior that guides the segmentation process. Finally, our work also contributes with a methodology that revives classical shape descriptor theory \cite[Chap. 11]{gonzalez2018digital}, once central to computer vision but largely eclipsed by deep learned features, as a means for improving the performances of vision algorithms.

% \sigrids integrate robust superpixel shape invariants with appearance and color cues, enabling a richer representation of image content. This combination of classical invariants and modern learning frameworks also helps explain \sigrid's effectiveness in image segmentation.

% It also should be noted that although this work evaluates \sigrid on binary image segmentation, it is intended as a general-purpose visual representation applicable to a broad class of spatial reasoning tasks, from image classification to object localization.

\section{Methodology}
\label{sec: methodology}

\subsection{Overview and Motivation}
% We introduce \sigrid (Superpixel-Integrated Grid), a structured representation that captures both spatial and semantic image information while substantially reducing input dimensionality. 
Conventional pixel-level representations store only local color values and redundantly encode similar information across neighboring pixels, resulting in large and often wasteful memory usage. Their regular grid structure, however, is highly compatible with convolution-based operations. In contrast, superpixels compress boundary and appearance cues into a compact set of feature vectors, but their irregular organization makes them difficult to integrate directly into standard architectures such as CNNs.

We introduce \sigrid (Superpixel-Integrated Grid) a structured image representation that leverages superpixels' perceptually coherent regions to construct a compact and meaningful input grid. Each cell in this grid stores descriptors derived from an entire superpixel, including both appearance and shape-based features. This enables downstream models to process more semantically enriched and spatially compressed inputs, reducing computational overhead while preserving critical visual cues. Therefore, \sigrid both retains the compressed semantic prowess and size of superpixels, while enabling their use in convolution-based networks, as Table \ref{tab:comp} summarizes.

\subsection{\sigrid Construction}

Figure \ref{fig:sigrid_construction}(a)-(d) illustrates the method to derive a \sigrid $S \in \mathbb{R}^{d \times w' \times h'}$ from an image $I \in \mathbb{R}^{c \times w \times h}$ of $c$ channels and its superpixelation $P \in \{1, \ldots, K\}^{w \times h}$ with $K$ superpixels. 

Our algorithm begins by computing the spatial center of each superpixel in $P$ and overlaying an $w' \times h'$ grid on top of $P$. The grid size $n$ is chosen as the smallest value such that no two centers fall within the same cell (with a small number of collisions allowed in practice; see Section~\ref{sec:experiments}). Each superpixel is then assigned to a grid location, and \sigrid $S$ is constructed by placing the corresponding $d$-dimensional descriptors \cite{hu1962visual} into their assigned non-empty cells, while leaving the remaining cells zero-filled. These descriptors, discussed in more detail in Section~\ref{sec:experiments}, encode both color and shape attributes of each superpixel, thus capturing the essential information in the associated image region. In terms of computational overhead, \sigrids can be quickly superpixel generated using fast GPU-friendly  algorithms \cite{fastslic_github} and descriptors computed with scattering techniques \cite{pytorch_scatter}.

\subsection{Network Training and Inference using \sigrids}

When training an image segmentation model using \sigrids, we first convert the training ground-truth segmentations into \sigrid-like objects, i.e. $n\times n$ matrices with segmentation labels on grid locations containing superpixel cells. The remaining grid locations are given ``empty'' labels, which signal to our optimizer that losses (cross-entropy in the case of segmentation) won't be computed for those tensor coordinates.

We assign a segmentation label to a superpixel based on the majority pixel-level label contained within its boundary. As superpixels closely delineate the natural object borders in the image, pixels within a given superpixel will most likely fall entirely within a single image region, and label ambiguities inside superpixels are rare.

At inference time, the model predicts an $w' \times h'$ grid of binary labels. These predictions are expanded back to pixel space by assigning each superpixel's pixels the predicted label of the \sigrid cell it was assigned to. Figure \ref{fig:sigrid_construction}(e)-(f) illustrate this process.

\subsection{Comparative Analysis}

\sigrid primitives provide a variety of technical and computational advantages to pixels and superpixels:
\setlist{nolistsep}
\begin{itemize}[noitemsep]
    \item They approximately preserve the spatial dispositions of superpixels, while arranging them on a regular grid.
    \item Because they are regular tensors, akin to images but smaller, they can be readily fed into standard CNNs while preserving their internal architecture.
    \item They make use of superpixel shape descriptors, which provide immediate structural image information to the network to process. Superxelations also provide natural edge guidance to segmentation networks.
    \item \sigrids are sparse by nature, which  leads to memory savings compare to pixels.
\end{itemize}

\section{Results and Discussion}\label{sec:experiments}

\begin{figure}
    \centering
    \includegraphics[width=1\linewidth]{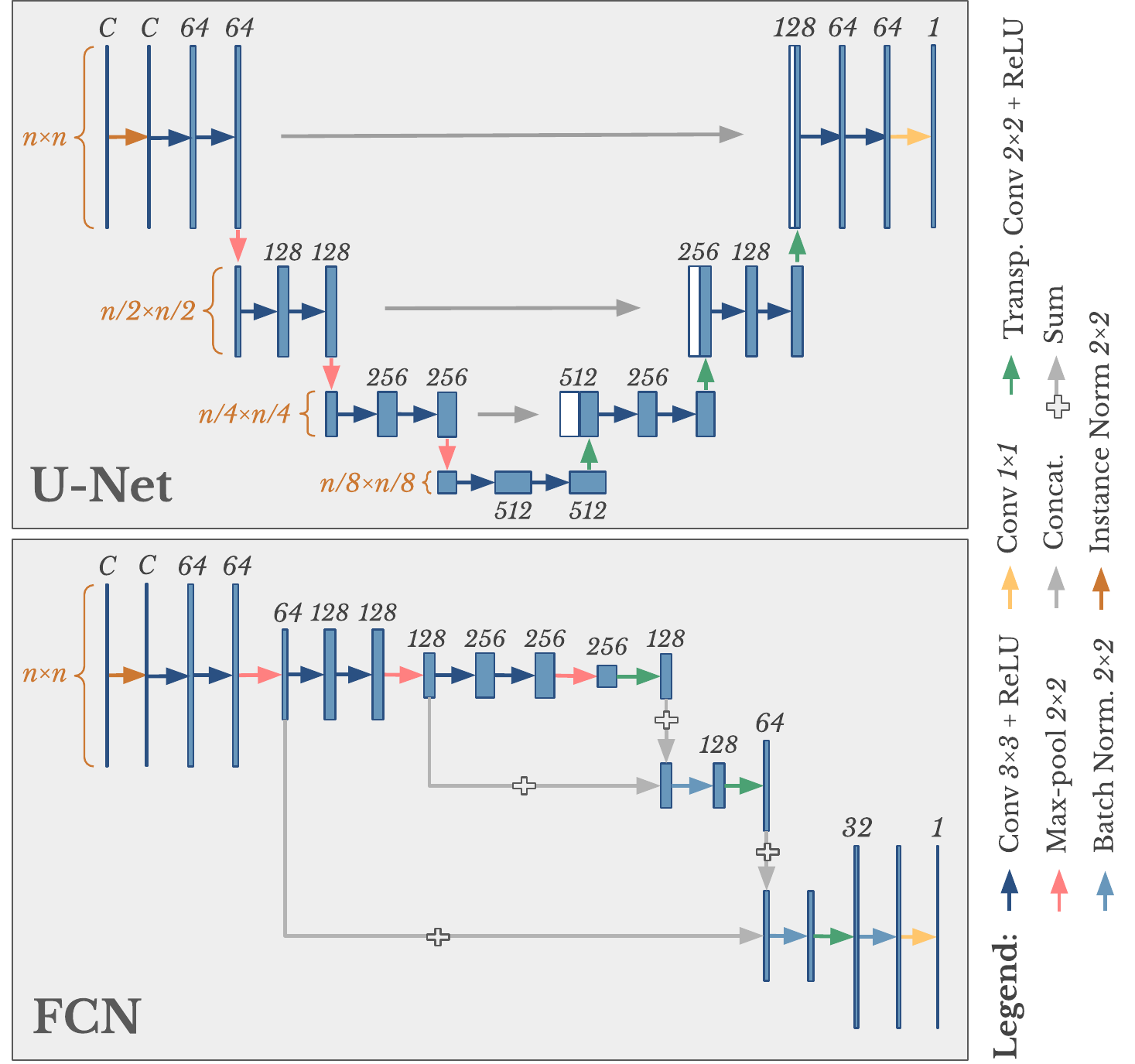}
    \vspace{-20pt}
    \caption{Architectural details of the CNNs in our experiments.}
    \label{fig:Archs}
\end{figure}

\begin{table}[!b]
    \caption{Effect of SLIC parameters on segmentation performance. 
    Reported values are Test Pixel \iou, with \maxiou in parentheses and the percentage of superpixel collisions in square brackets. 
    Experiments use U-Nets trained on CUB with \sigrids and average color descriptors only.}
    \vspace{3pt}
    \small
    \centering
    \setlength{\tabcolsep}{4pt}
    \resizebox{\linewidth}{!}{%
\begin{tabular}{lcccccc}
    \toprule
    & \multicolumn{6}{c}{\textbf{Number of Superpixels} ($K$)} \\
    \cmidrule(lr){2-7}
    \textbf{Compactness} & 500 & 700 & 1000 & 1500 & 2000 & 3000 \\
    \midrule

    \multirow{1}{*}{\textbf{5}}
        & 0.781 & 0.803 & 0.810 & 0.810 & 0.805 & 0.790 \\[-3pt]
        & {\scriptsize(0.915)} & {\scriptsize(0.924)} & {\scriptsize(0.931)} & {\scriptsize(0.936)} & {\scriptsize(0.936)} & {\scriptsize(0.920)} \\[-3pt]
        & {\scriptsize[0\%]} & {\scriptsize[0.05\%]} & {\scriptsize[0.13\%]} & {\scriptsize[0.36\%]} & {\scriptsize[0.73\%]} & {\scriptsize[2.07\%]} \\[2pt]

    \multirow{1}{*}{\textbf{10}}
        & 0.784 & 0.787 & 0.802 & 0.815 & 0.817 & 0.803 \\[-3pt]
        & {\scriptsize(0.922)} & {\scriptsize(0.929)} & {\scriptsize(0.935)} & {\scriptsize(0.939)} & {\scriptsize(0.939)} & {\scriptsize(0.924)} \\[-3pt]
        & {\scriptsize[0\%]} & {\scriptsize[0.03\%]} & {\scriptsize[0.08\%]} & {\scriptsize[0.26\%]} & {\scriptsize[0.59\%]} & {\scriptsize[2.03\%]} \\[2pt]

    \multirow{1}{*}{\textbf{20}}
        & 0.786 & 0.795 & 0.807 & \textbf{0.818} & 0.806 & 0.820 \\[-3pt]
        & {\scriptsize(0.925)} & {\scriptsize(0.931)} & {\scriptsize(0.937)} & {\textbf{\scriptsize(0.941)}} & {\scriptsize(0.942)} & {\scriptsize(0.932)} \\[-3pt]
        & {\scriptsize[0\%]} & {\scriptsize[0.05\%]} & {\scriptsize[0.05\%]} & {\textbf{\scriptsize[0.16\%]}} & {\scriptsize[0.42\%]} & {\scriptsize[1.92\%]} \\[2pt]

    \multirow{1}{*}{\textbf{100}}
        & 0.789 & 0.813 & 0.816 & 0.803 & 0.818 & 0.819 \\[-3pt]
        & {\scriptsize(0.913)} & {\scriptsize(0.923)} & {\scriptsize(0.930)} & {\scriptsize(0.938)} & {\scriptsize(0.940)} & {\scriptsize(0.941)} \\[-3pt]
        & {\scriptsize[0\%]} & {\scriptsize[0\%]} & {\scriptsize[0.01\%]} & {\scriptsize[0.02\%]} & {\scriptsize[0.12\%]} & {\scriptsize[1.35\%]} \\[2pt]

    \bottomrule
\end{tabular}
}
    \label{tab:superpixel_discarded}
\end{table}

\begin{table}[!hb]
\caption{Effect of superpixel descriptors choice. We test Average Color (AC), Area (A), Width (W), Height (H), Compactness (C), Solidity (S), Excentricity (E) and Hu Moments (Hu). \cmark\xspace indicates the feature was selected in that configuration. For this superpixel setting, $\operatorname{Max IoU} = 0.941$.}
\label{tab:original_unet_transposed}
\vspace{3pt}
\centering
\small
\setlength{\tabcolsep}{4pt}

\resizebox{\linewidth}{!}{%
\begin{tabular}{cccccccc cccc}
\toprule
\multicolumn{8}{c}{\textbf{Descriptors}} & \multicolumn{4}{c}{\textbf{Performance Metrics}} \\
\cmidrule(lr){1-8} \cmidrule(lr){9-12}
\textbf{AC} & \textbf{A} & \textbf{W} & \textbf{H} & \textbf{C} & \textbf{S} & \textbf{E} & \textbf{Hu} &
\textbf{\shortstack{Cell\\Acc.}} & \textbf{\shortstack{Pixel\\Acc.}} & \textbf{\shortstack{Cell \\ IoU}} & \textbf{\shortstack{Pixel \\ IoU}} \\
\midrule
\cmark & \xmark & \xmark & \xmark & \xmark & \xmark & \xmark & \xmark & 95.50 & 95.31 & 0.824 & 0.818 \\
\cmark & \cmark & \xmark & \xmark & \xmark & \xmark & \xmark & \xmark & 95.61 & 95.46 & 0.826 & 0.820 \\
\cmark & \xmark & \cmark & \cmark & \xmark & \xmark & \xmark & \xmark & 95.70 & 95.54 & 0.830 & 0.824 \\
\cmark & \cmark & \cmark & \cmark & \xmark & \xmark & \xmark & \xmark & 95.71 & 95.54 & 0.829 & 0.821 \\
\cmark & \xmark & \xmark & \xmark & \cmark & \cmark & \cmark & \xmark & 95.71 & 95.55 & 0.831 & 0.824 \\
\xmark & \xmark & \xmark & \xmark & \xmark & \xmark & \xmark & \cmark & 92.22 & 92.19 & 0.710 & 0.699 \\
\xmark & \xmark & \cmark & \cmark & \xmark & \xmark & \xmark & \cmark & 92.13 & 92.09 & 0.705 & 0.698 \\
\cmark & \xmark & \xmark & \xmark & \xmark & \xmark & \xmark & \cmark & \textbf{95.81} & \textbf{95.64} & \textbf{0.833} & \textbf{0.826} \\
\xmark & \xmark & \xmark & \xmark & \cmark & \cmark & \cmark & \cmark & 92.93 & 92.86 & 0.732 & 0.722 \\
\cmark & \cmark & \cmark & \cmark & \cmark & \cmark & \cmark & \cmark & 95.55 & 95.38 & 0.826 & 0.819 \\
\bottomrule
\end{tabular}%
}
\end{table}

\begin{table*}
    \caption{Comparison of Baseline and \sigrid variants for U-Net and FCN across CUB, DUTS, DUTS-OMRON, and ECSSD datasets. All results are for 100 training epochs. \sigrid models use $80 \times 80$ grids with average color and Hu moment features.}
    \label{tab:merged_sigrid_unet_fcn}
    \centering
    \scriptsize
    \setlength{\tabcolsep}{3pt}
    \vspace{3pt}
    \resizebox{\linewidth}{!}{%
    \begin{tabular}{lcccccccccccccccc}
        \toprule
        & \multicolumn{4}{c}{\textbf{CUB}} 
        & \multicolumn{4}{c}{\textbf{DUTS}} 
        & \multicolumn{4}{c}{\textbf{DUTS-OMRON}} 
        & \multicolumn{4}{c}{\textbf{ECSSD}} \\
        \cmidrule(lr){2-5} \cmidrule(lr){6-9} \cmidrule(lr){10-13} \cmidrule(lr){14-17}
        & \shortstack{U-Net \\ (Pixel)} & \shortstack{U-Net \\ (\sigrid)} & \shortstack{FCN \\ (Pixel)} & \shortstack{FCN \\ (\sigrid)}
        & \shortstack{U-Net \\ (Pixel)} & \shortstack{U-Net \\ (\sigrid)} & \shortstack{FCN \\ (Pixel)} & \shortstack{FCN \\ (\sigrid)}
        & \shortstack{U-Net \\ (Pixel)} & \shortstack{U-Net \\ (\sigrid)} & \shortstack{FCN \\ (Pixel)} & \shortstack{FCN \\ (\sigrid)}
        & \shortstack{U-Net \\ (Pixel)} & \shortstack{U-Net \\ (\sigrid)} & \shortstack{FCN \\ (Pixel)} & \shortstack{FCN \\ (\sigrid)} \\
        \midrule
        Pixel Acc. (\%)   & 96.791 & \textbf{95.641} & 93.847 & \textbf{93.678} 
                    & 89.543 & \textbf{88.281} & 83.626 & \textbf{85.122} 
                    & 90.326 & \textbf{91.096} & 88.614 & \textbf{88.908} 
                    & 83.833 & \textbf{83.093} & 79.876 & \textbf{80.666} \\
        Pixel \iou    & 0.871 & \textbf{0.826} & 0.786 & \textbf{0.768}
                    & 0.703 & \textbf{0.665} & 0.613 & \textbf{0.605}
                    & 0.690 & \textbf{0.682} & 0.604 & \textbf{0.625}
                    & 0.631 & \textbf{0.619} & 0.552 & \textbf{0.589} \\
        Pixel $\mathrm{MaxF}_\beta$
                    & 0.873 & \textbf{0.797} & 0.752 & \textbf{0.736}
                    & 0.626 & \textbf{0.592} & 0.484 & \textbf{0.488}
                    & 0.631 & \textbf{0.625} & 0.585 & \textbf{0.537}
                    & 0.670 & \textbf{0.629} & 0.568 & \textbf{0.563} \\
        Cell Acc. (\%)    & --      & \textbf{95.810} & --      & \textbf{93.759}
                    & --      & \textbf{88.520} & --      & \textbf{85.314}
                    & --      & \textbf{91.554} & --      & \textbf{89.347}
                    & --      & \textbf{83.420} & --      & \textbf{80.907} \\
        Cell \iou     & --      & \textbf{0.833} & --      & \textbf{0.765}
                    & --      & \textbf{0.664} & --      & \textbf{0.610}
                    & --      & \textbf{0.698} & --      & \textbf{0.636}
                    & --      & \textbf{0.617} & --      & \textbf{0.589} \\
        Cell $\mathrm{MaxF}_\beta$  
                    & -- & \textbf{0.807} & -- & \textbf{0.740}
                    & -- & \textbf{0.582} & -- & \textbf{0.490}
                    & -- & \textbf{0.668} & -- & \textbf{0.574}
                    & -- & \textbf{0.635} & -- & \textbf{0.569} \\
        \midrule
        \shortstack[l]{Time \\ (s/epoch)}   
                    & 89.58  & \textbf{26.97}  & 82.61  & \textbf{18.79}  
                    & 74.62  & \textbf{24.41}  & 63.72  & \textbf{15.79}  
                    & 28.92  & \textbf{9.69}   & 23.89  & \textbf{6.29}  
                    & 5.54   & \textbf{2.28}   & 4.60   & \textbf{1.56} \\
        GFlops  & 43.12  & \textbf{7.24}   & 12.55  & \textbf{2.14}   
                & 43.12  & \textbf{7.24}   & 12.55  & \textbf{2.14}   
                & 43.12  & \textbf{7.24}   & 12.55  & \textbf{2.14}   
                & 43.12  & \textbf{7.24}   & 12.55  & \textbf{2.14} \\
        \bottomrule
    \end{tabular}
    }

\end{table*}

\subsection{Implementation Remarks and Experimental Setup}

Each image is segmented with a GPU-optimized implementation of the SLIC algorithm \cite{achanta2012slic} \cite{fastslic_github}, using a specified number of segments $K$ and a compactness parameter that balances boundary adherence with region regularity. In practice, when $K$ is large, we allow for a minimal amount of superpixel center collisions, where two or more centroids share the same \sigrid. In this case, we discard the smaller superpixel in area from the pipeline. To address tightly clustered centers when computing \sigrids, we also merge superpixels whose centroids are closer than a resolution-dependent threshold $\tau = \max(h, w) / n$, where $h$ and $w$ are the image dimensions and $n$ is the target \sigrid resolution. Pairs with $d_{ij} < \tau$ are merged transitively, after which the segmentation is relabeled and centroids recomputed. 

We efficiently compute superpixel descriptors using GPU-backed tensor scattering operations \cite{pytorch_scatter}, which take
values from a source tensor (an image in our case) and accumulate them into a target tensor at specific indices (given by our superpixelation). We implemented and tested the following superpixel descriptors:
    \setlist{nolistsep}
    \begin{itemize}[noitemsep]
        \item \textit{Appearance-based:} Average RGB color (3 channels),
        \item \textit{Shape-based:} Superpixel area, width, height, compactness, eccentricity, solidity (1 channel each) \cite[Chap. 11]{gonzalez2018digital}, and Hu moments \cite{hu1962visual} (7 channels).
    \end{itemize}

We evaluate \sigrid on four standard segmentation benchmarks: CUB \cite{wahCUB2011}, DUT-OMRON \cite{yang2013saliency}, DUTS \cite{wang2017} and ECSSD \cite{shi2015hierarchical} using their official train/test splits. The images in these datasets have an average size of approximately $310 \times
375$ pixels. For both \sigrid and pixel-level training, we augmented the data with random rotation and horizontal/vertical flips. Augmentations are applied prior to any superpixel computation. We set $w' = h' = 80$ for all experiments.

We evaluate \sigrid using two popular convolutional architectures: the U-Net \cite{ronneberger2015u} and the Fully Convolutional Network (FCN) \cite{long2015fully}, whose designs follow the networks used in \cite{graham2017submanifold}. Figure \ref{fig:Archs} shows the implementation details of our tested networks. In both cases, the input of the model is first standardized using an Instance Normalization layer \cite{ulyanov2016instance}, which mitigates scale imbalances arising from its various descriptor values and ensures that they all contribute comparably.

We evaluate segmentation quality at both the \sigrid-cell and pixel levels using three standard metrics: Mean Intersection over Union (\iou), accuracy, and the $\mathrm{MaxF}_\beta$-score (with $\beta = 0.3$). RGB baselines, which process conventional $3 \times w \times h$ images, are evaluated only at the pixel level. For \sigrids, we additionally report \maxiou, defined as the pixel-level \iou that would be obtained if all non-empty \sigrid cells were perfectly classified. This quantity represents the upper bound on pixel-level performance achievable under a given superpixelation. Since superpixels may not align exactly with ground-truth boundaries, we typically have \maxiou$<1$.

All models were trained using binary cross-entropy, AdamW (learning rate $10^{-3}$, weight decay $10^{-4}$), for 100 epochs with batch size $32$, and implemented in Pytorch \cite{paszke2019pytorch}\footnote{The code used to generate the results can be found at \url{www.github.com/JackRobs25/SIGrid}} and trained on an NVIDIA A100 GPU with 40 GB of RAM.

% \subsection{Ablation Studies}

% We fine-tuned and evaluated our \sigrid method through two ablations on the CUB dataset and a comparison to pixel-based baselines across all four datasets. We observed that a \sigrid size of $80 \time 80$ (i.e., $n = 80$) struck a suitable balance between compactness and coverage of the superpixels across all datasets.

\subsection{Optimal SLIC Hyperparameters}
\label{subsec: SLIC}
To select the optimal SLIC hyperparameters (compactness and number of segments $K$), we tested various superpixel settings for \sigrids with just average color descriptors (Table~\ref{tab:superpixel_discarded}). We test U-Nets trained on CUB for these experiments and those of the next section. Increasing segments beyond 1500 yielded marginal \iou improvements but significantly increased discarded superpixels. Thus, we chose $K = 1500$ segments and $\operatorname{compactness} = 20$, balancing minimal superpixel loss (0.16\%) and strong segmentation accuracy (\iou: 81.82\%). The suitability of this choice across datasets was confirmed with DUTS, DUTS-OMRON, and ECSDD, losing just 0.046\%, 0.051\% and 0.060\% superpixels, respectively.

\subsection{Effective \sigrid Channel Configuration}
We next identified the most effective superpixel descriptors (Table~\ref{tab:original_unet_transposed}). Using average color alone achieved high performance (pixel \iou: 0.818). Incorporating geometric or shape features offered minimal gains on top, with the addition of Hu moments yielding the highest \iou of 0.826. This result elucidates the importance of aggregating both superpixel appearance and shape data for performance. Our final \sigrid setting, therefore, combines average color and Hu moments. 

\subsection{\sigrid vs. Pixel Baseline Results}
Table~\ref{tab:merged_sigrid_unet_fcn} compares \sigrid against pixel-based inputs for U-Net and FCN across the four datasets. \sigrid closely matched baseline segmentation performance, with \iou gaps typically under 0.05. Notably, \sigrid-FCN outperformed pixel-FCN on DUTS-OMRON and ECSSD (0.625 vs. 0.604 \iou and 0.589 vs. 0.552, resp.). Beyond accuracy, \sigrid delivers substantial efficiency gains. Training is up to $4\times$ faster, while computational cost (GFLOPs) is reduced by as much as $6\times$, highlighting a favorable trade-off between accuracy and efficiency. Preprocessing overhead remains negligible, with \sigrid construction taking under 0.1s per image across all datasets (specifically, CUB: 0.091s, DUTS: 0.074s, DUT-OMRON: 0.095s, ECSSD: 0.093s).

We attribute \sigrid's performance to two factors:
\setlist{nolistsep}
\begin{itemize}[noitemsep]
    \item Superpixels provide a strong edge guidance to segmentation, improving less complex networks such as FCN.
    \item Shape descriptors not only offset the detriment caused by the resolution reduction, but also give structural image data for simple models to predict the segmentation.
\end{itemize}

% \begin{table}[H]
%     \centering
%     \makebox[\textwidth]{
%     \begin{tabular}{lccccccc}
%         \hline
%         \multicolumn{1}{c}{} & \multicolumn{7}{c}{\textbf{n\_segments}} \\
%         \cline{2-8}
%         \textbf{Compactness} & \textbf{300} & \textbf{500} & \textbf{700} & \textbf{1000} & \textbf{1500} & \textbf{2000} & \textbf{3000} \\
%         \hline
%         \textbf{5}   & 62.00 () & 63.53 () & 62.83 () & 66.60 () &  68.28 ()       &         & 64.31 () \\
%         \textbf{10}  & 61.42 () & 64.38 () & 64.05 () &  66.24 ()   67.75    &         &         & 66.28 () \\
%         \textbf{20}  & 60.45 () & 65.07 () & 65.71 () &   67.71  ()    &         &         & 66.42 () \\
%         \textbf{100} & 60.56 () & 63.72 () & 66.15 () &    67.89 ()     &         &         & 69.36 () \\
%         \hline
%     \end{tabular}
%     }
%     \caption{Full U-Net 80x80 CUB (Binary IoU). Values are Test Pixel IoU (Maximum Test Pixel IoU).}
%     \label{tab:full_unet_cub_combined}
% \end{table}

% References should be produced using the bibtex program from suitable
% BiBTeX files (here: strings, refs, manuals). The IEEEbib.bst bibliography
% style file from IEEE produces unsorted bibliography list.
% -------------------------------------------------------------------------
\section{Conclusion and Future Work}
We introduce a novel image data primitive, \sigrid, which combines the semantic richness and compactness of superpixels with the regular structure required by convolutional networks for segmentation. Our results show that \sigrid achieves performance comparable to, and in some cases surpassing, pixel-level baselines, while vastly reducing training and inference time to a fraction of the cost. Looking ahead, we plan to exploit the inherent sparsity of \sigrid through sparse convolution operations to further accelerate CNN training \cite{graham2017submanifold} and extend these results to multiregion segmentation.

\newpage
\bibliographystyle{IEEEbib}
\bibliography{main}

\end{document}